\documentclass[preprint,times,review]{elsarticle}

\usepackage[utf8]{inputenc} % allow utf-8 input
\usepackage[T1]{fontenc}    % use 8-bit T1 fonts
\usepackage{hyperref}       % hyperlinks
\usepackage{url}            % simple URL typesetting
\usepackage{booktabs}       % professional-quality tables
\usepackage{amsfonts}       % blackboard math symbols
\usepackage{nicefrac}       % compact symbols for 1/2, etc.
\usepackage{microtype}      % microtypography
\usepackage{xcolor}         % colors
\usepackage{amsmath,xparse}
\usepackage{amssymb}
\usepackage{comment}
\usepackage{graphicx}
\usepackage{ulem}
\usepackage{multirow}       % for merging cells in table

\newcommand{\ap}{\textcolor[rgb]{0,0,1}} %% adeline's edits
\newcommand{\apn}[1]{\textcolor[rgb]{0,0.8,0.2}{[AP: {#1}]}} %% adeline's notes     - master of latex! -fr
\newcommand{\far}{\textcolor[rgb]{0,0.8,1}} %% felix's edits
\newcommand{\farn}[1]{\textcolor[rgb]{0.4,0.8,0.1}{[FR: {#1}]}} %% felix's notes

\newcommand{\E}{\mathrm{E}}
\newcommand{\Var}{\mathrm{Var}}

\DeclareMathOperator{\arcsinh}{arcsinh}

\setcitestyle{numbers,open={[},close={]}}

\begin{document}

\begin{frontmatter}

\title{Learnable Gabor modulated complex-valued networks for orientation robustness}

% -----------------------------------
% Authors
% -----------------------------------
\author[1]{Felix Richards\corref{cor1}}
\ead{felixarichards@gmail.com}

\author[2]{Adeline Paiement}
\ead{adeline.paiement@univ-tln.fr}

\author[1]{Xianghua Xie}
\ead{x.xie@swansea.ac.uk}

\author[3]{Elisabeth Sola}
\ead{elisabeth.sola@astro.unistra.fr}

\author[3]{Pierre-Alain Duc}
\ead{pierre-alain.duc@astro.unistra.fr}

% -----------------------------------
% Authors' addresses
% -----------------------------------
\cortext[cor1]{Corresponding author}

\address[1]{Department of Computer Science
 Swansea University
 Swansea, UK}
 
\address[2]{Université de Toulon, Aix Marseille Univ, CNRS, LIS
 Marseille, France}
 
\address[3]{Université de Strasbourg
 CNRS,
 Observatoire astronomique de Strasbourg,
 France}

\begin{abstract}
Robustness to transformation is desirable in many computer vision tasks, given that input data often exhibits pose variance. While translation invariance and equivariance is a documented phenomenon of CNNs, sensitivity to other transformations is typically encouraged through data augmentation. We investigate the modulation of complex valued convolutional weights with learned Gabor filters to enable orientation robustness. The resulting network can generate orientation dependent features free of interpolation with a single set of learnable rotation-governing parameters. By choosing to either retain or pool orientation channels, the choice of equivariance versus invariance can be directly controlled. Moreover, we introduce rotational weight-tying through a proposed cyclic Gabor convolution, further enabling generalisation over rotations.
We combine these innovations into Learnable Gabor Convolutional Networks (LGCNs), that are parameter-efficient and offer increased model complexity. We demonstrate their rotation invariance and equivariance on MNIST, BSD and a dataset of simulated and real astronomical images of Galactic cirri.
\end{abstract}

%Robustness to transformation is desirable in many computer vision tasks, given that input data often exhibits pose variance. While translation invariance and equivariance is a documented phenomenon of CNNs, sensitivity to other transformations is typically encouraged through data augmentation. We investigate the modulation of complex valued convolutional weights with learned Gabor filters to enable orientation robustness. The resulting network can generate orientation dependent features free of interpolation with a single set of learnable rotation-governing parameters. By choosing to either retain or pool orientation channels, the choice of equivariance versus invariance can be directly controlled. Moreover, we introduce rotational weight-tying through a proposed cyclic Gabor convolution, further enabling generalisation over rotations.
%We combine these innovations into Learnable Gabor Convolutional Networks (LGCNs), that are parameter-efficient and offer increased model complexity \sout{while keeping backpropagation simple}. We demonstrate their rotation invariance and equivariance on MNIST, \sout{Berkeley Segmentation Dataset (BSD)} \far{BSD} and a dataset of simulated and real astronomical images of Galactic cirri \sout{for segmentation and denoising of textured contaminant regions}.
% for segmentation and denoising of textured contaminant regions


\end{frontmatter}
%---------------------------
\section{Introduction}
\label{sec:intro}

%\sout{possess multiple structural symmetries, with objects exhibiting a range of pose variation}
We enable learning  of approximate orientation invariance and equivariance in convolutional neural networks (CNN).
Datasets in various domains often exhibit a range of pose variation (e.g.~scale, translation, orientation, reflection).
CNNs are inherently equipped to handle translation invariance, but remedies for other symmetries often involve large models and datasets with plenty of augmentation.
This inability to properly adapt to transformations such as local/global rotations is a major limitation in CNNs.

An important distinction is that of equivariance versus invariance. For a network to be equivariant, it should be robust to variation in pose and be able to carry over transformations of the input to transformed features and output.
For tasks where output is dependent on these transformations, network invariance alone is suboptimal as transformation information is discarded, by definition.
For example, in ultra deep astronomical imaging, the scattered light from foreground Galactic cirrus contaminates and occludes interesting Low-Surface Brightness (LSB) extragalactic objects.
%low surface brightness (ULSB) telescope imaging, cirrus clouds exhibiting clear orientation dependent features often contaminate and occlude interesting objects.
These cirrus clouds exhibit orientation dependent features:
segmenting cloud regions is a problem requiring invariance, as orientation of cloud streaks does not necessarily affect the geometry of the cloud's envelope.
On the other hand, removing occluding clouds, which is crucial to studying background LSB galaxies, is a denoising problem that requires robust and descriptive equivariant features.

Numerous works have been published alongside CNN research attempting to integrate forms of rotation invariant and equivariant feature learning in an a-priori fashion.
Approaches typically generate rotation dependent responses by one of the following strategies: 1) learning orientations by constructing filters from a steerable basis \cite{Cohen2017SteerableCnns, Worrall, Weiler}, 2) rotating convolution filters/input by preset angles \cite{Marcos2017RotationNetworks, Marcos2016LearningClassification, Laptev2016TI-POOLING:Networks}, or 3) introducing orientation information through analytical filters \cite{Luan2018GaborNetworks,Zhou2017OrientedNetworks}.
A significant drawback of the former type is that it introduces significant computational overhead \cite{Cheng2018RotDCF:Networks}.
In the second category, the rotation process imposes the use of interpolation which results in artefacts for any rotation outside of the discrete sampling grid.
This is overcome in the latter category by using analytical filters with an inherent rotation parameter. Orientations are static in \cite{Luan2018GaborNetworks,Zhou2017OrientedNetworks}, similarly to the second category, however there is no inherent limitation of analytical filters preventing them from having learnable orientation parameters.
There is thus a need for a dynamic orientation sensitive architecture that can accurately adapt to the input's transformation. We address this need in this work using Gabor filters, analytical filters that are parameterised by orientation, scale and frequency among other variables.
Furthermore, Gabor filters are differentiable with respect to their parameters, meaning that these parameters can be learned through steepest descent style algorithms.

\begin{comment}
\sout{Steerability overcomes this but requires that filters are explicitly constrained to be constructed from a basis of preset filters} \apn{it might be a good idea to talk about the fixed Gabor filter network to illustrate what you mean}, \sout{reducing expressivity \far{in the case where strict equivariance is not optimal}} \apn{by 'optimal', so you mean 'attainable with the limited range of available filters'?} \farn{Yes}. 
%This is also arguably at a much higher implementation cost \farn{Is this a thing?} \apn{it depends. do you mean 'complexity in coding', or 'computational cost' i.e. 'requiring lots of memory or computation time'?} \farn{coding complexity} \apn{I have rarely seen that presented as a strong limitation} \farn{I thought so}.
\sout{Thus there is a need \far{for an orientation sensitive architecture that can learn the optimal level of equivariance depending on the dataset}.} \farn{Trying to say the opposite of filters that are not constrained - without being vague, and that orientations are not fixed} \apn{let's see how the previous comments are resolved first, that may help...}
\end{comment}

\textbf{Contributions - }In this paper we propose Learnable Gabor Convolutional Networks (LGCN), a complex-valued CNN architecture highly sensitive to rotation transformations. We utilise adjustable Gabor modulation of convolutional weights to generate dynamic orientation activations.
%\farn{Think of experiment to demonstrate this 'dynamacity'} \apn{you could use fixed orientations as a baseline to compare against in all experiments}
By learning Gabor parameters alongside convolutional filters we achieve features that are dependent on exact angles with no interpolation artefacts.
Moreover, there is no explicit constraint on convolutional filters, allowing a diverse feature space that adapts to the degree of rotation equivariance required.
We extend the modulation approach used in \cite{Luan2018GaborNetworks} to complex space, enabling use of the full complex Gabor filter and exploiting the inherent descriptive power of complex neurons.
Further, we build on this and propose a convolutional operator where Gabor filter modulation is cyclically shifted, inspired by group theory CNNs \cite{Cohen2016, Cohen2017SteerableCnns, Dieleman2016, Bekkers2018Roto-translationAnalysis, Weiler}, allowing propagation of orientation information throughout a forward pass in an equivariant manner.

%\apn{is this (and learnable orientations) the only differences between the two methods? If there is more, it would be good to list clearly, at some point, what are all the differences to the most related works. I am under the impression that you forget to talk about the cyclic process} \farn{Has this been sufficiently addressed?} \apn{The differences have been addressed yes. How about listing the cyclic aspect here, since you are listing all the innovative (use of) concepts and their benefit for invariance and equivariance?} \farn{yes of course, it's easy to glance over these old paragraphs}.

\begin{comment}
For example, object classification typically does not require equivariance as categorisation should remain constant under any deformity of the object, i.e. invariance is sufficient \farn{wordy}. In the case of object segmentation, any deformity in the object's geometry should often carry over into the final segmentation geometry: this is a case where transformation equivariance is beneficial. \apn{I like this comparative definition of invariance and equivariance, it is helpful to the readers who are not familiar with the distinction between the two.} \apn{but I prefer the one of the previous paragraph because it is more directly related to your experiments} \farn{+1}
\end{comment}

%---------------------------
\section{Previous works}
\label{sec:previous}

Methods have been developed in an attempt to integrate transformation invariance in an a-priori fashion. Prior to CNN popularity, the use of hand-crafted features such as SIFT \cite{Lowe1999ObjectFeatures} and Gabor filters \cite{Haley1999Rotation-invariantModelb, Arivazhagan2006TextureFeatures} was explored to generate rotation/scale invariant representations.
A widely adopted technique in deep learning is to augment transformations into a dataset \cite{Simard2003BestAnalysis., Krizhevsky2012, Ciregan2012Multi-columnClassification}. This brute force approach introduces new samples to prompt the model to learn this new range of transformations. Models with learned invariance through augmentation require a very large parameter space to capitalise on data augmentation, and still may generalise poorly to unseen transformations.

% Some recent works encode symmetries into CNN architectures.
There has been much work recently on encoding symmetries into CNN architectures. 
Early efforts utilised pooling over transformed responses, e.g. siamese networks \cite{Wu2015DeepAuto-annotation}, training-time augmentation \cite{Kanazawa2014LocallyNetworks,Sohn2012LearningTransformations}, parallel convolutional layers \cite{Dieleman2014,Dieleman2016}, kernel-based affine pooling \cite{Gens2014DeepNetworks}, and image warping \cite{Henriques2017WarpedTransformations, Jaderberg2015SpatialNetworks, Lin2017InverseNetworks, Jia2016DynamicNetworks}.
Specifically in the last few years there has been a surge of interest in rotation equivariant architectures.
Authors have been able to formulate CNNs entirely from principles of group theory and thus construct modified operators and/or constrain filters \cite{Cohen2016, Cohen2017SteerableCnns, Cohen2018SphericalCnns}.
\citet{Bekkers2018Roto-translationAnalysis} employ bi-linear interpolation to enable any regular sampling of the continuous group of 2D rotation. Similarly \cite{Marcos2016LearningClassification} and \cite{Marcos2017RotationNetworks} utilise copied and rotated filters, but pool over the produced activation maps. While interpolation allows rotation by exact angles (as in \cite{Bekkers2018Roto-translationAnalysis, Marcos2016LearningClassification, Marcos2017RotationNetworks}) it introduces artefacts for angles outside of the discrete sampling grid. In \cite{Jacobsen2017DynamicNetworks} residual blocks are combined with principles of steerable bases to learn approximate equivariance. \citet{Worrall} allow exact orientation representations while overcoming dependence on interpolation by constraining filters to the family of complex circular harmonics: there is a clear demonstration of complex neurons encoding rotational information which justifies our usage of complex CNNs for rotation equivariance. \citet{Finzi2020GeneralizingData} are able to construct group equivariance without steerable filters by constructing filters as parameterisations of Lie algebra .
Similarly, \citet{Weiler} present a CNN architecture with learnable steerable filters, and derive a generalised weight initialisation method for steerable basis coefficients. Using a formulation of steerable filter architectures, \cite{Weiler2019GeneralCNNs} proposes a general framework for equivariant networks under any combination of rotation, reflection or translation. 
We draw inspiration from the cyclic shifting group convolutions commonly used in group theory based CNNs \cite{Cohen2016, Cohen2017SteerableCnns, Dieleman2016, Bekkers2018Roto-translationAnalysis, Weiler}, and propose a similar operation for rotation generalisation without requiring derivation from group theory and reducing computational overhead.

Analytical filters have made a resurgence in many deep learning contexts. Specific to transformation invariance, analytical filters parameterised by rotation are fast and can extract orientational features dependent on exact angles, overcoming interpolation artefacts. Several approaches replace convolutional weights with wavelet filters \cite{Sifre2013RotationDiscrimination,Bruna2013InvariantNetworks, Ejbali2018AClassification}. Wavelets are also applied to inputs of standard convolutional layers in a preprocessing fashion \cite{Fujieda2018WaveletNetworks}. In \cite{Wang2018ModulatedNetworks} authors present a framework for convolutional weight modulation, achieving enhanced filters with binarised weights. Zhou et al. \cite{Zhou2017OrientedNetworks} exploit rotation parameterisation of discrete Fourier transforms to extract orientation information, modulating standard convolutional filters with a filter bank of rotated analytical filters. Luan et al. \cite{Luan2018GaborNetworks} implement a similar approach but opt to use Gabor filters, demonstrating that they are more robust to rotation and scale transformations. In \cite{Khan2018LearningWavelets} wavelet filter hyperparameters are learned in an end to end fashion for spectral decomposition through wavelet deconvolutions. We combine lessons learned from \cite{Khan2018LearningWavelets} with \cite{Luan2018GaborNetworks} to construct Gabor filter modulation with learnable parameters.

%---------------------------
\section{Methodology}
\label{sec:method}

\begin{figure*}[t]
  \begin{center}
    \includegraphics[width=1\textwidth]{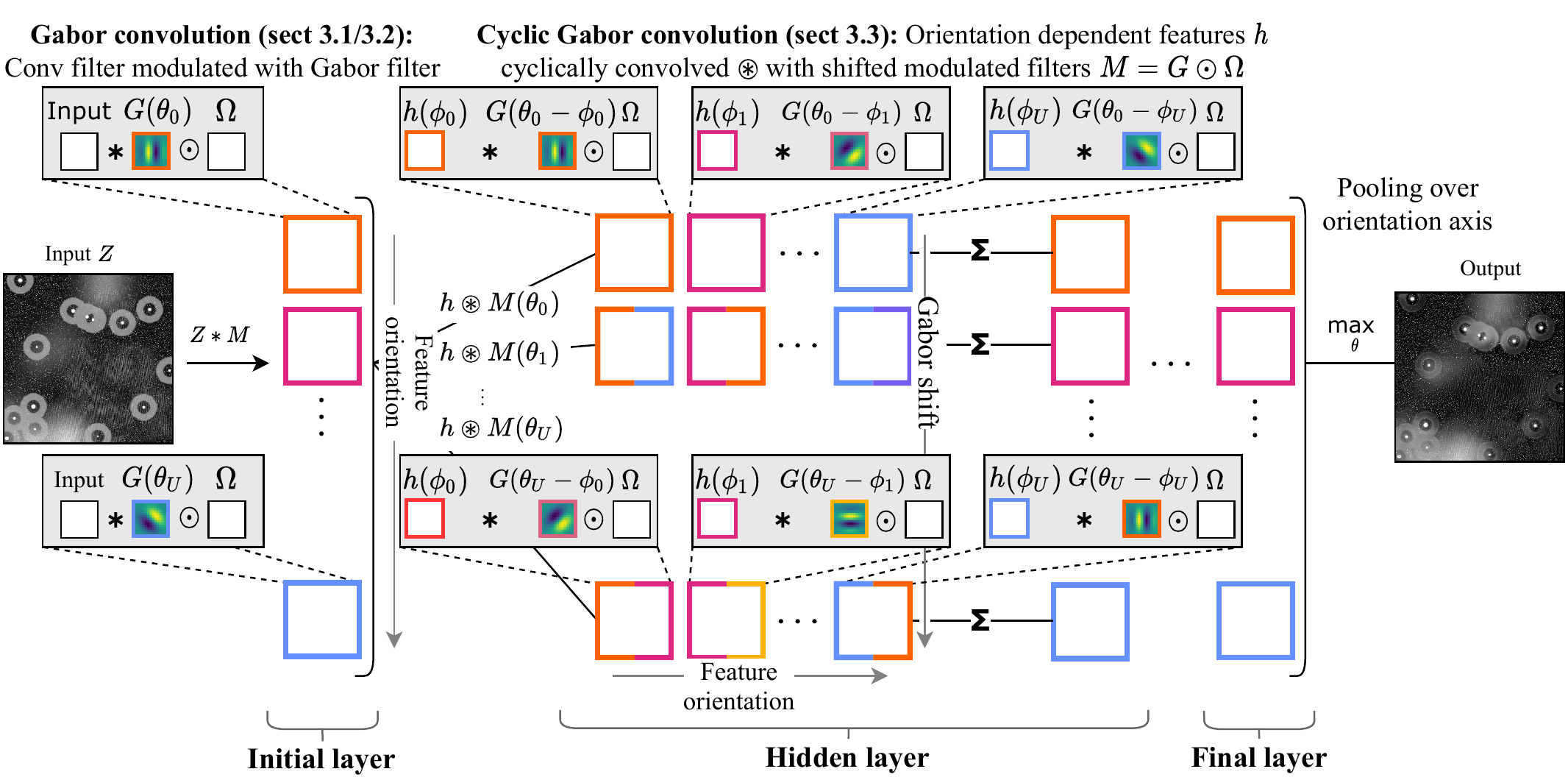}
  \end{center}
  \caption{Overview of LGCN, with illustration of the filter modulation and cyclic convolution concepts. Input channel indices are omitted to highlight the rotational aspect of the proposed method. Colour denotes individual feature orientations. To achieve orientation invariance, an additional pooling operation (not represented here) may be added at the end of each hidden layer to pool over orientations. In this equivariant problem example, orientation information is retained across layers. (For interpretation of the references to colour in this figure caption, the reader is referred to the web version of this article.)} 
  \label{fig:general}
\end{figure*}

%In the invariant case, orientation pooling can be applied \ap{over the feature orientation dimension (for convolution layers) or the Gabor shift dimension (for cyclic convolution layers)} \sout{following} \ap{at the end of} each hidden layer \sout{in addition to the final layer} \apn{This may be too much details for a caption, but it would be good to put this description in the main text, maybe in section 3.4?}. \farn{I was about to say! }

Our LGCN achieves sensitivity to rotation transformations through adjustable Gabor modulation of convolutional weights. In the architecture defined below, modulation parameters are learned alongside convolutional filters.
Having separate modulation and convolution parameters keeps backpropagation simple. Given that convolution filters are not explicitly constrained as in other methods attempting to overcome transformations, the result is a larger space of possible features.
Fig.~\ref{fig:general} illustrates the general structure of LGCNs, providing an overview of the concepts proposed throughout this section.
%\sout{, and \sout{adds no explicit regularisation on} \ap{does not explicitly constrain or limit} how convolutional filters learn} \apn{If I understood correctly, the second half of this sentence is redundant with the next one and could be removed to save space}.
An important development of this approach is that parameters belong to complex space, allowing both real and imaginary parts of analytical filters to be utilised. Given that frequency response filters are often designed over complex space, this enables a variety of modulation choices. %\sout{This is compounded by the inherent ability of complex phase to encode rich geometric information \cite{Oppenheim1981TheSignals} such as orientation, making LGCNs vastly equipped to handle rotation transformations.}

LGCNs can consider several orientations simultaneously, which are finely tuned to the task being solved.
LGCNs are able to achieve activations dependent on arbitrary continuous rotations with no interpolation artifacts and without using steerable filter bases.
With modulation, LGCNs increase model complexity at little cost to the parameter size. In our case Gabor filters are calculated with orientation $\theta$ and wavelength $\lambda$, meaning they can generate a new feature channel with only two parameters. Finally, LGCNs utilise a novel convolutional operator where Gabor filters are cyclically shifted during modulation, enabling propagation of orientation information across layers and thus facilitating learned invariance and equivariance.
%\apn{is there no advantage for invariance?}

%\sout{First we present a framework for generalised modulation of complex valued convolutional filters. Though in this paper we choose to modulate with Gabor filters, it is possible to modulate with a variety of complex analytical filters with this approach. Following this, we detail the exact forward and backward pass processes for the spatial Gabor filter. We then present cyclic Gabor convolutions which introduce rotational weight sharing into convolutional filters. Finally, we provide some practical discussion on implementing Gabor modulation in CNNs.}

\subsection{Filter Modulation of Complex-Valued Networks}
\label{subsec:modulation}

In order to enable compatibility with a wide range of analytic frequency response filters, we construct complex-valued CNN layers.
As described in \cite{Trabelsi2018DeepNetworks} we construct complex numbers by encoding real and imaginary parts as separate real valued elements. A complex convolutional weight tensor can be written as $\Omega = A + iB$, where $A$ and $B$ are stored internally as real tensors. Complex arithmetic is then simulated with appropriate real operations acting on these elements. For a complex valued input $H = X + iY$, convolution is computed as:
\begin{align}
    Z = \Omega \ast H = (A \ast X - B \ast Y) + i(B \ast X + A \ast Y)
\end{align}
where $\ast$ denotes the convolution operator.
For nonlinearities, we use the complex ReLU proposed in \cite{Arjovsky2016UnitaryNetworks}, $\mathbb{C}\text{-ReLU}(Z) = \text{ReLU}(|Z|+b)\frac{Z}{|Z|}$ with $b$ a real-valued bias term.
We also implement complex analogues of batch normalisation, given in \cite{Trabelsi2018DeepNetworks}, and average spatial pooling, trivially given by considering the average of real and imaginary parts separately.

For an analytical filter $\Phi{^P}=\Phi^{{P}}_\text{Re}+i\Phi^{{P}}_\text{Im}$ with $D$ parameters $p_d \in P=\{ p_0,...,p_{D-1} \}$ we extend the convolutional modulation presented in \cite{Wang2018ModulatedNetworks} to complex space.
The modulation of convolutional kernel $\Omega^c=A^c+iB^c$ of channel $c$ with discretised filter $\Phi^{P}$ is given by $M^{c,P} = \Phi^{P} \odot \Omega^c$, where $\odot$ represents complex element-wise multiplication.
Output from convolution with the modulated filter is then given by $Z^{c,P} = M^{c,P} \ast H$, or for each pixel at coordinates ($s$, $t$):
\begin{align}
    z_{s,t}^{c,P}&= \sum\limits_{k=1}^K \sum\limits_{l=1}^K m_{k,l}^{c,P} \ h_{s+k,t+l}
\end{align}

This construction of modulated filters can be viewed as a collection of filter banks, where both the underlying kernels (via convolutional filter $\Omega$) and frequency sub-bands (via $\Phi$) are learnt. The complete filter bank has dimensions $2\times C_\text{out} \times C_\text{in} \times U \times K \times K$, with $C_\text{out}$ and $C_\text{in}$  the number of output and input convolutional channels respectively, $U$ the number of modulating filters, and $K$ the convolution kernel size. Any given filter is obtained by modulating a convolutional filter $W^c$ of channel $c$ with analytical filter $\Phi^{P^u}$. A significant advantage of this formulation is that a filter bank of $U$ filters is created from a single canonical filter, meaning that encoding of transformation representations requires little computational overhead. Adjusting parameters through backpropagation requires calculating the gradient of a differentiable loss function $L$ with respect to $p_d$:
\begin{align}
    \frac{\partial L}{\partial p_d} &= \frac{\partial L}{\partial M^{c,P}} \frac{\partial M^{c,P}}{\partial p_d} 
    = \sum\limits_{k=1}^K \sum\limits_{l=1}^K \frac{\partial L}{\partial m_{k,l}^{c,P}} \frac{\partial m_{k,l}^{c,P}}{\partial p_d}
%    = \sum\limits_{k=1}^K \sum\limits_{l=1}^K \bigg( \sum\limits_{s=1}^N \sum\limits_{t=1}^N \frac{\partial L}{\partial z_{s,t}^{c,P}} \frac{\partial z_{s,t}^{c,P}}{\partial m_{k,l}^{c,P}} \bigg) \frac{\partial m_{k,l}^{c,P}}{\partial p_d} \\
    &= \sum\limits_{k=1}^K \sum\limits_{l=1}^K \frac{\partial m_{k,l}^{c,P}}{\partial p_d} \sum\limits_{s=1}^N \sum\limits_{t=1}^N \frac{\partial L}{\partial z_{s,t}^{c,P}} h_{s+k,t+l} .
\end{align}

Thus the only constraint on choice of analytical filter $\Phi$ is that it is differentiable with respect to parameter $p_d$. In the following subsection we compute the above derivative in the scenario where only a subset of parameters are learned.

\subsection{Learnable Gabor filters}
\label{subsec:gabor}

In this paper we modulate with Gabor filters similarly to \cite{Luan2018GaborNetworks}, which are feature detectors characterised by rotation sensitivity and frequency localisation: $G(\lambda, \theta, \psi, \sigma, \gamma)_{k,l} = e^{-\frac{k'^2+\gamma^2l'^2}{2\sigma^2}} e^{i(\frac{2\pi}{\lambda} k' + \psi)}$ with $k'=k\cos\theta+l\sin\theta$ and $l'=l\cos\theta-k\sin\theta$.
Two major differences with \cite{Luan2018GaborNetworks} is that we work with complex-valued networks, and we learn the parameters of the filters while they were fixed to static orientations in \cite{Luan2018GaborNetworks}.
A significant advantage of Gabor filters in comparison to Fourier related methods such as DCT is that they are not constructed from a sinusoidal basis, meaning that discontinuous patterns, such as edges, can more easily be represented. We fix (hyper)parameters other than orientation $\theta$ and wavelength $\lambda$: $G(\lambda, \theta, 0, \frac{1}{\sqrt{2}}, 1)$ as in \cite{Luan2018GaborNetworks} which demonstrated that this provides sufficient expressivity while simplifying computation.
Though we choose to modulate with the well-documented Gabor filters due to orientation and frequency parameterisation, it is possible to modulate with a variety of complex analytical filters with this approach.

Thus the modulated filter $M^{c,P}$ can be written as $M^{c,P} = G^{P} \odot \Omega^c$. We evaluate ${\partial m_{k,l}^{c,P}}/{\partial p_d}$ at pixel $k,l$ in the context of Gabor filter modulation for both parameters. Given that $\theta, \lambda \in \mathbb{R}$ we treat $M^{c,P}$ as a function of the real and imaginary parts separately ($k,l$ indices omitted for readability):
% \begin{align}
%     \frac{\partial m_{k,l}^{c,P}}{\partial \theta} &= {a'}_{k,l}^c \frac{\partial G_\text{Re}}{\partial \theta} + {b'}_{k,l}^c \frac{\partial G_\text{Im}}{\partial \theta}
%     = \frac{2\pi}{\lambda} e^{-(k^2+l^2)} l'[-{a'}_{k,l}^c\sin(\frac{2\pi}{\lambda}k') + {b'}_{k,l}^c\cos(\frac{2\pi}{\lambda}k')]\\
%     \frac{\partial m_{k,l}^{c,P}}{\partial \lambda} &= {a'}_{k,l}^c \frac{\partial G_\text{Re}}{\partial \lambda} + {b'}_{k,l}^c \frac{\partial G_\text{Im}}{\partial \lambda}
%     = \frac{2\pi}{\lambda^2} e^{-(k^2+l^2)} k' [{a'}_{k,l}^c\sin(\frac{2\pi}{\lambda}k') - {b'}_{k,l}^c\cos(\frac{2\pi}{\lambda}k')],
% \end{align}
\begin{align}
    \frac{\partial m^{c,P}}{\partial \theta} &= {a'}^c \frac{\partial G_\text{Re}}{\partial \theta} + {b'}^c \frac{\partial G_\text{Im}}{\partial \theta}
    = \frac{2\pi}{\lambda} e^{-(k^2+l^2)} l'[-{a'}^c\sin(\frac{2\pi}{\lambda}k') + {b'}^c\cos(\frac{2\pi}{\lambda}k')]\\
    \frac{\partial m^{c,P}}{\partial \lambda} &= {a'}^c \frac{\partial G_\text{Re}}{\partial \lambda} + {b'}^c \frac{\partial G_\text{Im}}{\partial \lambda}
    = \frac{2\pi}{\lambda^2} e^{-(k^2+l^2)} k' [{a'}^c\sin(\frac{2\pi}{\lambda}k') - {b'}^c\cos(\frac{2\pi}{\lambda}k')],
\end{align}
where ${a'}^c = a^c + b^c$, ${b'}^c = a^c - b^c$.
Backpropagation $\frac{\partial L}{\partial p_d} = \frac{\partial L}{\partial M^{c,P}} \frac{\partial M^{c,P}}{\partial p_d}$ can now be calculated, enabling learning of Gabor filters' parameters alongside convolutional weights. Accordingly, parameters are updated by $\theta'=\theta -\eta\frac{\partial L}{\partial \theta}$ and $\lambda'=\lambda-\eta\frac{\partial L}{\partial \lambda}$, with $\eta$ denoting learning rate.
%\farn{Summarise subsection and lead onto next.}
%Given a standard 2D convolutional layer with kernels of size $k \times k$, $F_\text{in}$ input channels and $F_\text{out}$ output channels, an equivalent complex convolutional layer would have weight dimensions $2 \times F_\text{out} \times F_\text{in} \times k \times k $.

\subsection{Cyclic Gabor Convolutions}
\label{subsec:cyclic}

In intermediate layers we implement cyclic convolutions to further increase rotation information without increasing parameter size and utilise the additional feature channels generated by Gabor modulation.
We exploit the cyclic property of finite subgroups of 2D rotation transformations to create convolutional filters based on all permutations of orientation and canonical filters.
By sharing all weights across every orientation, the underlying canonical filters further generalise over rotations.
This is analogous to how filters are exposed to all translations in standard CNNs to encourage generalisation over translations.
This propagation of rotation dependence directly facilitates equivariance, in contrast to non-cyclic Gabor convolutions which must pool over the orientation axis per layer.
% \sout{The cyclic filter shift} \apn{we had both versions un-commented, so I'm not sure of what happened}

Note that this cyclic framework does not require analytical filters that are steerable, only that filters can be parameterised by rotation.
That is to say, it is not a requirement that filters meet the criteria of linear steerability according to \citet{Freeman1991TheFilters}.
In particular, we demonstrate that rotational weight sharing through cyclic shifting can be achieved with Gabor filters, which are not steerable. Networks constructed from steerable filters where basis coefficients are learned in place of convolutional kernels inherently and explicitly limit the filter space -- whether this is a downside or an optimal regularisation to achieve rotation equivariance is yet to be shown. In comparison, learnable modulation with Gabor filters implicitly regularises filter space.

The cyclic convolution design we propose takes inspiration from group convolutions presented in \cite{Cohen2016}.
Specifically, cyclic Gabor convolutions utilise the shifting operation used in group convolutions defined over 2D roto-translations \cite{Cohen2016, Weiler}.
As the modulation transformation cannot be used to form a symmetry group, we do not derive computation using the group framework.
However, using the orientation sensitivity of the Gabor filter we implement a similar resulting feature composition, enabling rotational weight sharing without requiring a proof for strict equivariance.
%In comparison to the steerable filters learned through group convolutions \cite{Weiler}, \farn{comment on structural differences}.
%We are also able to gain rotational weight sharing without requiring a proof for strict equivariance.

With $h^c(\theta)$ denoting channel $c$ and orientation $\theta$ of the previous layer's activation map, for a single orientation and output channel, cyclic convolution $\circledast$ is computed as:
\begin{align}
    z^{\hat{c}} (\theta) &= \sum^{C_\text{in}}_{c=1} \Big{[} h^c \circledast  M^{\hat{c}c} \Big{]} (\theta)
    %&= \sum^{C_\text{in}}_{c=1} \sum_{\phi \in P} \Big{[} h^c (\phi) \ast  M^{\hat{c}c} (\theta - \phi) \Big{]} \\
    = \sum^{C_\text{in}}_{c=1} \sum_{\phi \in P} \Big{[} h^c (\phi) \ast \big{(} G (\theta - \phi) \odot \Omega^{\hat{c}c} \big{)} \Big{]}. \label{eq:group_derivation}
\end{align}
$P$ is the set of $U$ orientations that are used to generate Gabor filters.
Note this formulation allows any size of $P$.
In order to keep implementation efficient and avoid recalculating Gabor filters for all permutations of learned orientations we keep $\phi \in P$ as the original angles.
This choice allows filters to be reused with a cyclic shift of the orientation components per different output orientation $\theta$. 
\begin{comment}
\farn{Generalise convolutions to transformation groups other than translation.}
\farn{Group convolutions are defined with a transformation operation, which they are then equivariant under, i.e. $MATHS$.}
\farn{Strict equivariance cannot be proved with this formulation but we demonstrate that it aids with enabling approximate equivariance.}

\begin{align}
    \lambda' = \frac{2\pi}{\lambda}
\end{align}

We define the group's transformation operator $\rho$ as a complex rotation,

\begin{align}
    \rho_\theta \Phi = e^{-i\lambda' k' (\theta)} \Phi
\end{align}

Computation of the group convolutions follows similarly to \cite{Weiler} with the difference that the sinusoidal component of Gabor filters is not linearly steerable.

Substituting the Gabor filters into equation (\ref{eq:group_derivation}) shows that the rotation operator can be interpreted as acting only on the sinusoidal component of the Gabor filter

\begin{align}
    z^{\hat{c}c} (\theta) &= \sum^{C_\text{in}}_{c=1} \sum_{\phi \in P} \Big{[} h^c \ast e^{-i \lambda' k' (\phi)} \sum_{k,l} e^{i \lambda' k'(\theta-\phi)} e^{-(k^2+l^2)} \Omega^{\hat{c}c}_{k,l} \Big{]} \\
    &= \sum^{C_\text{in}}_{c=1} \sum_{\phi \in P} \Big{[} h^c \ast \sum_{k,l} e^{i \lambda' (k'(\theta-\phi) - k' (\phi))} e^{-(k^2+l^2)} \Omega^{\hat{c}c}_{k,l} \Big{]}
\end{align}

\end{comment}

\subsection{Learnable Gabor Convolutional Networks}
\label{subsec:lgcn}

The framework presented above allows learnable modulation to be added into any convolutional layer, making the method very versatile. There are some considerations to take into account however, which we discuss in this section.

\textbf{Complex weight initialisation - }
LGCNs operate over complex space, requiring weight initialisation to be rethought. Principles of He weight initialisation \cite{He2015DelvingClassification} no longer hold given that $\Var(\Omega) \neq \Var(A) +i\Var(B)$, i.e.~real and imaginary parts cannot be initialised independently. We use Trabelsi's generalisation of He's strategy over complex space \cite{Trabelsi2018DeepNetworks}, setting $\Var[|\Omega|]=\frac{4-\pi}{2n_{\text{in}}}$ with $n_\text{in}$ denoting the number of input units. The phase is then uniformly distributed around the circle. It is worth noting that He's derivation is specific to the traditional ReLU, using the result that for a given input $X_l$ to a layer $l$, and previous output $Y_{l-1}$: $\E[X^2_l]=\frac{1}{2}\Var[Y_{l-1}]$.
%\begin{align}
%    \E[X^2_l]=\frac{1}{2}\Var[Y_{l-1}].
%\end{align}
This holds for traditional ReLU, $X_l = \max(0, Y_{l-1})$, as $Y_{l-1}$ has zero mean and a symmetric distribution which is essentially split along its axis of symmetry. However with $\mathbb{C}$-ReLU, for $b<0$,
$Y_{l-1}$ is no longer divided along the axis of symmetry. For this reason we simply initialise the biases of $\mathbb{C}$-ReLU layers to zero. %\apn{was this already done in [37]?} \farn{They do not mention their initialisation strategy. Nor do they make any comments on the impact of relu biases on weight initialisation}

The choice of initialisation for modulation parameters is largely dependent on the choice of analytical filter, and should be influenced by the function's domain and the roles of individual variables. For initialisation of Gabor parameters, as discussed in Section \ref{subsec:gabor}, we fix phase shift $\psi$, aspect ratio $\gamma$ and scale $\sigma$ in order to simplify computation. Given that wavelength is a non-negative quantity we initialise $\lambda$ with mean $3\sqrt{U}$ and variance $\frac{\sqrt U}{4}$ as per \cite{He2015DelvingClassification}, and verified that training is stable. This choice of initialisation also avoids spatial aliasing of the Gabor filter for all kernel sizes (i.e. $3 \times 3$ or larger) at network initialisation. As the filter is sampled more than twice per phase, the signal is adequately captured, as per the Nyquist-Shannon sampling theorem. For orientation $\theta$, in the real case there is no benefit of using the full interval of rotations due to evenness, however in the complex case the oddness of the imaginary part causes orthogonal filters for $\theta$ with differing sign. For this reason we initialise $\theta$ uniformly around the full circle.%\farn{Diagram showing orthogonal gabor filters may be useful here} \apn{I agree} 

\textbf{Gabor axis considerations - }
Though the ability to create enhanced filters from a single canonical filter has advantages of parameter efficiency and weight-tying, it leaves the network prone to dimensionality explosion.
This can be controlled using one or more of three approaches depending on the problem at hand: adjusting the number of convolutional channels $C$ depending on the dataset's feature complexity; adjusting the number of modulating filters $U$ based on the dataset's pose variation; and max pooling along the orientation axis i.e.~over the modulating filters for each pixel of the bank of modulated feature maps.
The latter operation has the additional advantage of focusing the attention of the network on (local) dominant orientations, which is a particularly useful feature for orientation invariance.

\textbf{Invariance vs equivariance -}
There is a clear relation between pooling technique and invariance versus equivariance.
Preserving only the strongest orientation response discards low response representations and disentangles features, this is however at the cost of encouraging invariance to local rotations rather than equivariance.
In practice, invariance is achieved through pooling after each hidden layer over the feature orientation dimension or the Gabor shift dimension, for convolutions and cyclic convolutions, respectively -- see Fig.~\ref{fig:general}.
%In practice, invariance is achieved through pooling over the feature orientation dimension (for convolution layers) or the Gabor shift dimension (for cyclic convolution layers) at the end of each hidden layer -- see Fig.~\ref{fig:general}.
%\farn{diagram to illustrate this may be helpful} \apn{if we lack of space, this can be omitted. I find the explanation clear enough.} %\sout{For MNIST, max pooling over modulation is a suitable choice as the task only requires invariance of classification over rotation.} \apn{this sentence could be moved to the experiment section}

\textbf{Projection between $\mathbb{C}$ and $\mathbb{R}$ - }Finally, since data used in this paper is real, we set the imaginary part of inputs to zero. Some works \cite{Trabelsi2018DeepNetworks,Scardapane2018Complex-valuedFunctions} opt to include a preprocessing step to estimate the imaginary part though we found this had a detrimental effect on performance. For real classification, final complex feature maps must be projected back onto real space. We experimented with several projection methods such as complex linear layers and using magnitudes, but found that simply concatenating real and imaginary values into fully connected linear layers performed best. %\farn{should we include discussion of choice of projection? is this comparison something we want to present results on? e.g. concatenation of real/imaginary vs magnitude vs concatenation of mags/phases} \apn{If there is enough space to present the comparative results, then perfect. Otherwise, what you wrote is good, and the comparative results can be presented in the sup. materials as a bonus.}

%---------------------------
\section{Experiments}
\label{sec:results}

%\apn{The experiments section is hugely important and should be carefully structured to help interpreting the results. It is often a good idea to have one subsection/experiment per thing you want to demonstrate or compare.}
%\apn{The section could start by a summary of the aspects/properties you demonstrate/compare, in which scenarios, and the list of baseline and state-of-the-art methods that you use to compare.}

In this section we validate our learnable modulation formulation, showing that learning analytical filter parameters leads to improved accuracy on both artificial and real data. Initially, LGCNs are evaluated on variants of MNIST \cite{LeCun1998} containing rotated samples, where we evaluate the network's learned invariance. In the next section we compare invariance and equivariance in both a standard CNN and a learnable Gabor modulated CNN, where networks process synthesised samples of galactic cirri. All experiments throughout this section were run using a single NVIDIA GTX 1080 Ti.

\subsection{Orientation invariance on MNIST}
\label{subsec:mnist}

MNIST \cite{LeCun1998} (CC BY-SA 3.0 license)
is a standard benchmark for transformation invariance because of its simplicity, interpretability and vast array of variants.
We apply a random rotation between $[0, 2\pi)$ to yield a rotated MNIST, and train with 5-fold validation.
Our baseline classification architecture is similar to that used in \cite{Weiler, Cohen2016, Worrall}, with three blocks of increasing channels, representing a hierarchy of feature complexity. Each block contains two learnable Gabor modulated convolutional layers with a kernel size of $3 \times 3$ followed by max pooling along the orientation axis and average spatial pooling. We use no cyclic Gabor convolutional layers, but these may be included in future experiments. In the final block, features are pooled globally so that a given activation contains one complex value per feature channel. We then concatenate real and imaginary parts into a single vector and use three (real valued) fully connected layers for classification. The Adam optimiser \cite{Kingma2014Adam:Optimization} is used for network training, starting at a learning rate of 0.001 and then decaying with an exponential schedule by 0.9 every epoch. L2 weight regularisation is also enforced with a penalty of $10^{-7}$.
%Rotated MNIST \apn{if this is not MNIST-rot, make sure that people don't think that it might be, maybe by saying explicitly 'we rotate'} contains samples of handwritten digits randomly rotated between $[0, 2\pi)$.
%\sout{This is made significantly more challenging than the original rotated MNIST by restricting the training, validation and testing sets to} \sout{\ap{It contains} 10000, 2000 and 50000 \ap{training, validation and testing} samples, \sout{respectively} \ap{thus being smaller than the original MNIST}.} \apn{if it is a publicly available dataset, people can find its size easily} \farn{These details are included in lots of the cited papers} \apn{let's see how much space we have at the end... but it is not the most important thing to say} \farn{of course :)}

%\subsubsection{Orientation Robustness}

\begin{figure}[t]
  \begin{center}
    \includegraphics[width=\textwidth]{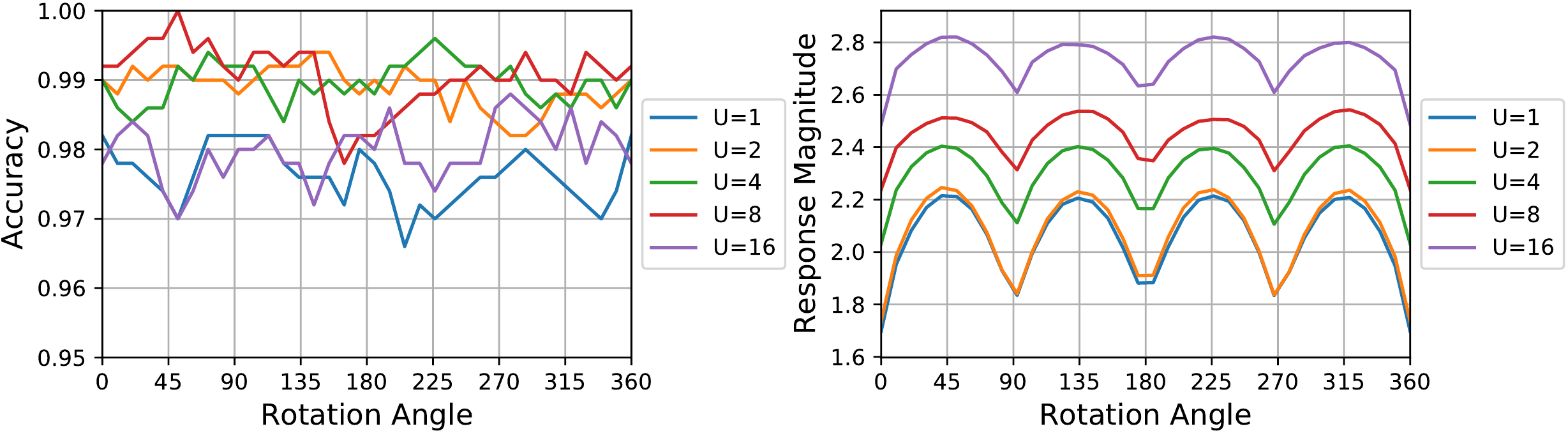} %[width=0.8\textwidth]
  \end{center}
  \caption{Effect of input rotation on MNIST classification accuracy (left) and magnitude of activations in the first modulated layer of the network (right), for different numbers of modulating filters and orientations, and on a subset of 1000 testing samples of MNIST.}
  \label{fig:mnist-orientation}
\end{figure}

\textbf{Exploration of rotation invariance in the feature maps --} The number of modulation filters $U$ has a direct effect on the network's ability to capture rotation dependent features. We vary this parameter and investigate its effect on network's performance and the learned features of the first layer, which are the most directly affected by low-level geometrical transformations of the input image. For this first experiment, we train networks with $U\in\{1,2,4,8,16\}$. We measure and compare response magnitudes (measured as the ratio between the average magnitudes of input and output activations) between original and rotated samples, for all rotations in the (discrete) range [0,$360^\circ$], for each network (Fig.~\ref{fig:mnist-orientation} right). 
Though response magnitude varies slightly, this may be largely due to interpolation artefacts caused by rotation of the input samples. Nonetheless, the pattern remains predictable throughout the rotation interval with decreasing amplitude for increasing $U$, indicating that the number of modulating filters has a direct impact on rotation invariance.
%demonstrating the stability of invariant features provided by using multiple orientation dependent modulation channels.}
%Throughout the rotation interval response magnitude remains, demonstrating the stability of invariant features produced by modulation.
%\farn{Some comments about the stability of features with rotation}
We also measure classification accuracy as a function of rotation for 1000 samples from the MNIST test set for each network (Fig.~\ref{fig:mnist-orientation} left). The small difference in accuracy between $U=1$ and others indicates that even a little orientation information is helpful in generating intra-class rotation-invariant features that remain inter-class separable. At $U=16$ there is a detrimental saturation of orientations possibly due to the model becoming too complex for the dataset size and task. Optimal performances are reached for $U$ between 2 and 8, with LGCN being not very sensitive to the exact value of this hyperparameter.

\textbf{Evaluation of the individual modifications to \cite{Luan2018GaborNetworks} --} We evaluate the performance improvements from our two modifications to \cite{Luan2018GaborNetworks} individually, namely the use of complex-valued filters and of learnable Gabor orientation parameters. In these experiments we apply these modifications both in turn and jointly to the model of \cite{Luan2018GaborNetworks}. The channel sizes for each LGCN variant were adjusted so that the total parameter size is at most equal to all of the compared models: for complex models this required halving the number of feature channels. The final results, shown in Tab.~\ref{tab:rotatedmnist}, show that both modifications improve classification accuracy, demonstrating the additional feature expressivity afforded in comparison to standard CNNs. All variants also outperform GCNs which use real and static Gabor modulation, with an absolute error difference of 0.6 for LGCNs, showing the benefit of our method's changes over the previous work. The combination of modifications leads to a large performance increase that may indicate a synergy between the two approaches. One possible explanation for this is that the complex Gabor filter provides smoother gradients with respect to $\theta$ and $\lambda$, as opposed to only the real part. We will test this hypothesis in future work.
%  \sout{To highlight the impact of learnable complex modulation \sout{and provide fair comparison to \cite{Luan2018GaborNetworks, Zhou2017OrientedNetworks},} \apn{the reviewers may argue that this would not be unfair, since it is just a method that you propose} we use no cyclic Gabor convolutional layers \apn{I'm not sure that this argument will be judged valid, since it is an ablation study: its purpose is to evaluate each modification in turn} but these may be included in future experiments.}

\begin{table}
\small
\caption{Classification accuracy on randomly rotated MNIST images.}
\begin{center}
\begin{tabular}{llllll}
\hline
%\multicolumn{3}{c}{LGCN (\ap{Proposed})}             &      &      \\
LGCN (proposed) & Complex static & Real learnable & GCN4 \cite{Luan2018GaborNetworks} & ORN8 \cite{Zhou2017OrientedNetworks} &  CNN \\ \hline
\textbf{0.9950}     & 0.9915        & 0.9911        & 0.9890           & 0.9888 &   0.9718 \\ \hline
\end{tabular}
\end{center}
\label{tab:rotatedmnist}
\end{table}

%\begin{itemize}
%    \item include diagram on filters/activataion maps to provide illustration for discussion of results. perhaps diagram with conv filters, filters and activation maps.
%    \item response magnitude plots across $U$: we can show the activation stability over rotations for different degrees of modulation. t-sne illustrate interclass feature separation better than response magnitude
%\end{itemize}

\textbf{Effect of adjusting learning strategies of Gabor parameters --} We investigate the effect of learning Gabor parameters other than $\theta$, and study how changing initialisation methods impacts the model's performance.
The default training configuration for LGCNs uses fixed $\sigma$ set at $\pi$, and learnable $\lambda$ initialised using a normal distribution with mean $3\sqrt U$ and variance $\frac{\sqrt U}{4}$.
We experiment with fixing or learning wavelength $\lambda$ and scale $\sigma$ and study how different configurations affect LGCNs.
In addition we compare additional weight initialisation strategies for these variables.
For wavelength initialisation we apply: fixed $\lambda=3$; normal distribution with unit mean and unit variance (not adjusting for $U$); and uniform distribution between $\lbrack -1.5,1.5\rbrack$.
For scale, in addition to fixed $\sigma=\pi$, we initialise with a normal distribution with mean equal to $\pi$ and quarter variance, and enable backpropagation.
Finally, we repeat these experiments with only one $\lambda, \sigma$ for all modulating Gabor filters per layer.

We train parameter restricted models with varying Gabor parameter learning strategies on rotated MNIST for 30 epochs with 5-fold validation, and record the average performance over all splits. Results are shown in Table \ref{tab:gaborparams}. While initialising with a normal distribution $\lambda = \mathcal{N}(3\sqrt{U}, \frac{\sqrt{U}}{4})$ and fixing $\sigma = \pi$ achieves the highest average performance, there is no clear strategy for either variable that remains best with the other variable strategy changed. Notably, there is a performance decrease when aliasing of the modulating Gabor filters is forcibly introduced by initialising wavelength $\lambda $ from a uniform distribution with bounds $\lbrack -1.5,1.5\rbrack$. In further tests it was noticed that in this scenario $\lambda$ values do not recover from this range of aliasing even after training for >100 epochs. For this experiment we conclude that given parameters are not in an aliasing range, LGCNs are not particularly sensitive to learning strategy of wavelength $\lambda$ and scale $\sigma$.

\begin{table}[]
\small
\caption{Classification accuracy on rotated MNIST averaged over 5 splits for different learning strategies of Gabor parameters wavelength $\lambda$ and scale $\sigma$. Rows are divided in the centre to denote whether a single $\lambda$ and $\sigma$ is used for all $U$ modulating Gabor filters, or $\lambda$ and $\sigma$ are separate for each modulating Gabor filter.}
\begin{tabular}{rlrrrr}
\hline
                                            &                                                               & $\lambda = \mathcal{U}(-1.5, 1.5)$ & Fixed $\lambda = 3$ & $\mathcal{N}(3, \frac{1}{4})$ & \multicolumn{1}{l}{$\mathcal{N}(3\sqrt{U}, \frac{\sqrt{U}}{4})$} \\ \hline
\multirow{2}{*}{Separate $\lambda, \sigma$} & Fixed $\sigma = \pi$                                          & 0.9672                             & 0.9702              & 0.9686                                         & 0.9692                                                                    \\
                                            & \multicolumn{1}{r}{$\sigma = \mathcal{N}(\pi, \frac{1}{4})$} & 0.9690                             & 0.9704              & 0.9693                                         & 0.9678                                                                    \\ \hline \hline
\multirow{2}{*}{Single $\lambda, \sigma$}   & Fixed $\sigma = \pi$                                          & 0.9684                             & 0.9707              & 0.9698                                         & \textbf{0.9713}                                                                    \\
                                            & \multicolumn{1}{r}{$\sigma = \mathcal{N}(\pi, \frac{1}{4})$} & 0.9673                             & 0.9699              & 0.9707                                         & 0.9685                                                                    \\ \hline                                                              
\end{tabular}
\label{tab:gaborparams}
\end{table}

\subsection{Invariance and equivariance to the dominant orientation of galactic cirri}
\label{subsec:cirrus}

\begin{figure}[h]
  \begin{center}
    \includegraphics[width=\textwidth]{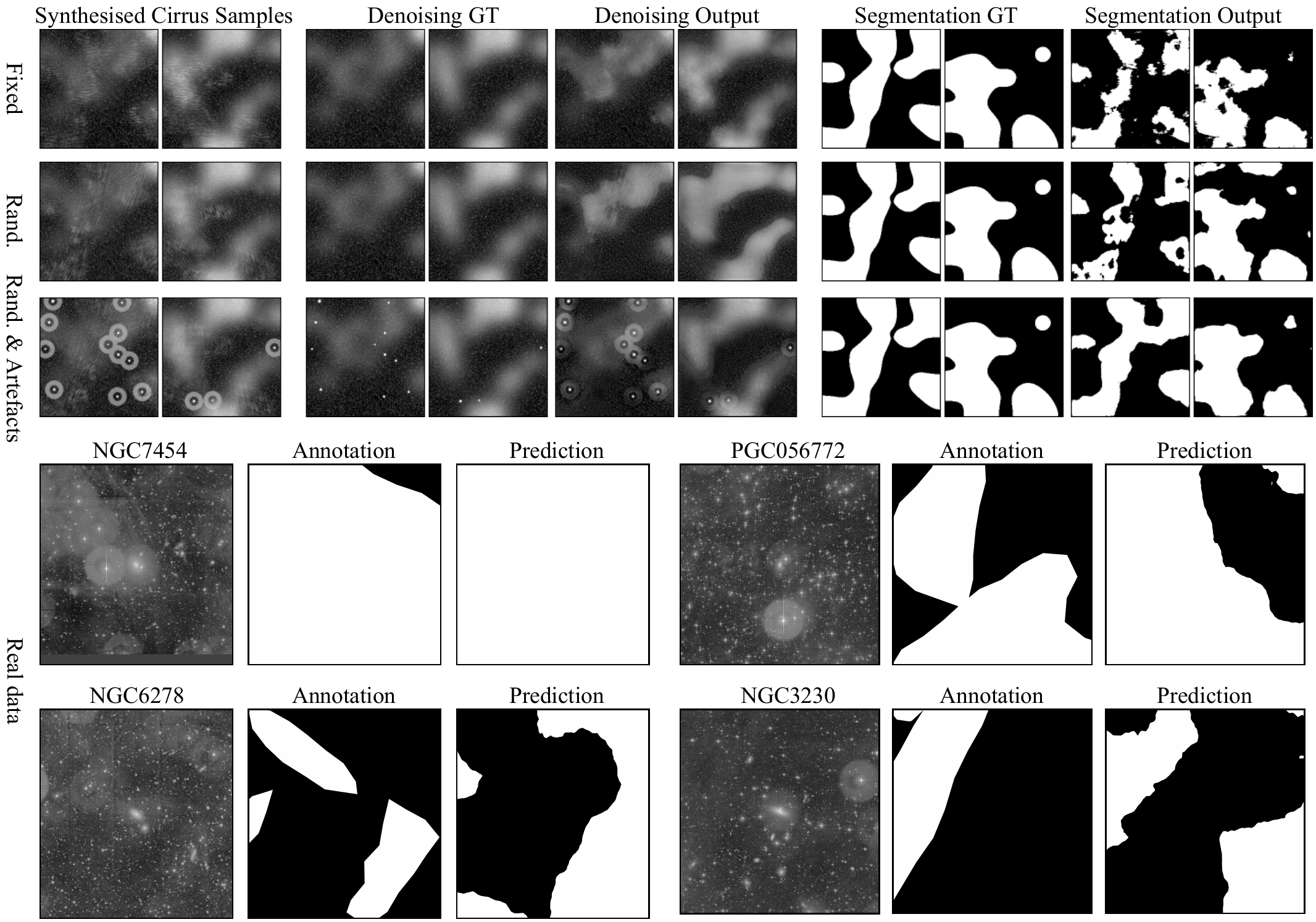} %[width=0.8\textwidth]
  \end{center}
  \caption{Denoising and segmentation results on real and synthesised samples of galactic cirri generated with fixed rotation; randomised rotation; and randomised rotation with stars and telescope artefacts. These are difficult tasks as the striped textures of cirrus regions are easily confused/obstructed with bright diffuse regions and other objects.}
  \label{fig:cirrus}
\end{figure}

We validate the benefit of modulation by applying LGCNs to a domain demanding robust orientation-sensitive features. We demonstrate that modulation not only enables the network to learn invariance and equivariance, but aids the network's ability to generate features unaffected by local disturbances. For these experiments we analyse samples of galactic cirrus clouds -- astronomical objects with striped quasi-textures exhibiting clear dominant orientations, as shown in Fig.~\ref{fig:cirrus} -- as they allow the design of experiments that assess both orientation invariance and equivariance separately and the comparative robustness between different models. These images are very challenging, exhibiting overlapping semi-transparent objects, including foreground cirrus with oriented patterns, background objects (e.g. galaxies) with vastly different textures and intensities, and telescope artefacts.
In the initial experiment we evaluate performance on various datasets composed from synthesised images of cirrus structures. Following this, we extend to a real world problem using low surface brightness telescope images containing cirrus contamination.
%Additional results are illustrated in the sup. materials.
%\apn{You could add or move here a very short discussion on the fact that this is a challenging dataset exhibiting overlapping semi-transparent objects, with foreground cirrus having dominant orientation, and background objects such as galaxies having completely different structures and textures.} 

\textbf{Generation of synthesised cirrus images --} In order to create images exhibiting discriminative features similar to real images of galactic cirrus, multiple noise patterns are combined. All synthesised images are formed from at least three parts, background $B$, cirrus $C$, and bright regions $R$. Pixels to form the background $B$ are drawn from a Gaussian distribution and inverted. 
%\apn{This reads as if the background was only Gaussian noise, but there are bright regions too, aren't there?} \farn{These are the bright regions, I have not documented how the point sources are created, I have just stated in the next para that they are added.} 
The cirrus image $C$ contains textured cloud shapes with smooth boundaries. The cloud shapes are produced by a 2D Gaussian mixture model (GMM) with 4 to 6 randomly located components with random standard deviations. The cirrus texture is created by combining a cloud texture and a streak texture, both generated from Perlin gradient noise of varying frequencies. 
A binary mask extracted from the GMM that forms the cirrus segmentation target.
Bright regions $R$ are smooth isotropic bright regions resembling regions of diffuse light, and are created from a GMM with a similar process to the cirrus case.
Finally, all parts are combined according to $\gamma B + \gamma C + R$ with the denoising target set to $\gamma B + R$,
before normalisation between 0 and 1,
where $\gamma=0.4$ balances bright regions vs. background and cirrus regions.

We design the dataset to have three variations of increasing realness. The first variation possesses only cirrus clouds with constant orientation and bright regions; the second randomises cirrus orientation; finally the third introduces star-like objects with telescope halo artefacts (i.e. bright transparent halos around each bright spots simulating stars). These star-like objects are created from a sharp Gaussian profile approximating a point source, where the standard deviation of each star's Gaussian profile is randomly slightly varied to ensure variation. A synthetic halo resembling a telescope artefact is then added around each star, and is created from a circle of fixed radius and width and with a uniform brightness proportional to the star's associated Gaussian standard deviation.
%\apn{is that the bright spots? Or are bright spots introduced only in the third dataset?}

Our synthesised dataset contains 300 samples: 160 for training, 40 for validation (for 5-fold validation) and 100 for testing.
%\apn{this paragraph doesn't need to be very long and detailed, but I agree with Elisabeth that it would be good to provide the main idea. In addition, it would be a good introduction to the analysis of the experimental results, if you explain briefly here what are the challenges of each dataset.}

\textbf{Comparing LGCN against a traditional CNN on synthesised cirrus images --} We create a U-Net \cite{Ronneberger2015U-net:Segmentation} style architecture in both standard form and with Gabor modulated convolutional layers, where skip connections are combined via summation (as in \cite{Quan2021FusionNet:Connectomics}) rather than concatenation. To enable comparison, we create four variants of this network: one with plain convolutions; one with complex-valued convolutions, denoted $\mathbb C$-CNN; one with static real Gabor filter modulation as in \cite{Luan2018GaborNetworks}; and one with learnable complex Gabor modulation with cyclic convolutions.
These networks are tasked with first segmenting the cirrus clouds, and secondly removing clouds and artefacts (if applicable). The complex filters of $\mathbb C$-CNN and LGCN naturally require twice the convolutional filter parameters. We ensure a fair comparison by adjusting channel sizes accordingly, thus keeping total parameter size of the two networks roughly equal. For the denoising task we do not utilise orientation pooling so that orientation information is preserved and equivariance is encouraged, as per discussion in \ref{subsec:lgcn}, and experimental verification. Results are presented in Tab.~\ref{tab:denoising_results} with IoU metric for segmentation and peak signal to noise ratio (PSNR) for denoising.
%\apn{It would be great to also have examples on real images. Maybe you could save space and at the same time include real examples by keeping only one row per scenario (and include more examples in the sup. materials to compensate)?}
%
\begin{table}
\small
\caption{Segmentation IoU (left), denoising PSNR (middle) on synthesised cirri with fixed and randomised orientation, and with stars and telescope artefacts. Segmentation IoU (right) on real cirrus samples in LSB images. *Gabor convolutions of \cite{Luan2018GaborNetworks} applied  to our base model}
\begin{center}
\begin{tabular}{l|ccc|ccc|c}
    \hline
     & Fixed & Rand. & \begin{tabular}[c]{@{}l@{}}Rand \&\\ artefacts\end{tabular} & Fixed & Rand. & \begin{tabular}[c]{@{}l@{}}Rand \&\\ artefacts\end{tabular} & Real cirrus \\ \hline
    Base model & 0.914 & 0.882   & 0.806                & 26.3  & 25.3    & 23.5          & 0.684        \\
    $\mathbb{C}$ base model  & 0.918 & 0.905 & 0.839 & 26.7 & 25.6 & 24.5 & 0.691 \\
    GC model*  & 0.923 & \textbf{0.920} & 0.875 & 26.7 & 26.2 & 24.8 & 0.685 \\
    %LGCN non-cyclic & - & - & - & - & - & - & - \\
    LGCN & \textbf{0.925} & 0.918   & \textbf{0.898}              & \textbf{27.4} & \textbf{27.1}    &\textbf{25.4}    & \textbf{0.731}\\
    \hline
\end{tabular}
\end{center}
\label{tab:denoising_results}
\end{table}

In the segmentation case, given that rotation of cirrus texture does not affect the cloud's envelope, this is a problem where invariance is beneficial. The denoising problem requires equivariance, as isolating the cloud from the detailed background is dependent on the orientation of its streaks.
In the first dataset, CNN, $\mathbb C$-CNN and GCN performances are close to LGCN's, generating fine segmentations of the cirrus clouds with few missed regions, and similarly good denoising. For the second, with orientation variation, GCN performs marginally best in the segmentation case but its performance drops for denoising. On the other hand, LGCN maintains a quite stable performance for denoising (and also for segmentation) on this dataset.
This difference in behaviour may be explained by the use of cyclic convolutions in LGCN that better preserve orientation information due to rotational weight-tying across layers.
CNN and $\mathbb C$-CNN performances start to fall behind, with relative differences of $4.1\%$ and $1.4\%$ IoU, respectively, for segmentation and $7.1\%$ and $5.9\%$ PSNR for denoising. This separation becomes much larger in the final experiment on the most complex data exhibiting overlapping textured regions and localised objects, with LGCNs outperforming GCNs in both tasks by $2.6\%$ and $2.4\%$, $\mathbb C$-CNN by $7.0\%$ and $3.7\%$, and plain CNNs by $11.4\%$ and $8.1\%$. The affect of randomising rotations and even introducing telescope artefacts makes little difference to LGCN's performance for segmentation, demonstrating its strength in generating rotation invariant features that are robust to local disturbances. We see that denoising performance is stable with randomised rotation, indicating equivariant encoding produced by the modulated layers. While performance drops for the third dataset, due to artefacts introducing strong variations locally, LGCN still outperforms other models by a larger margin than without artefacts, showing that feature robustness is exhibited in the equivariant case. The results demonstrate that both the use of complex numbers and the modulation of filters are beneficial. We note that the margin between CNN and $\mathbb C$-CNN is significantly the largest on this dataset variation requiring robustness, compared to other dataset variations. LGCN combines both augmentations and cyclic convolutions for further improved results. 
%\apn{Maybe this would be better demonstrated by also adding results of complex modulated filters with no cyclic conv.? Or am I getting confused?} \farn{It would remove the speculation but will take a week to implement and run}
Visual analysis of the network outputs (see Fig.~\ref{fig:cirrus}) indicates a possible overfitting for the first two datasets with no telescope artefacts for both the segmentation and denoising tasks, which results in a more difficult generalisation and poorer (visual) quality on test data. This issue may be due to these two simpler scenarios requiring simpler models and/or fewer training steps, and it will be investigated in future work.
%\apn{This last sentence could be slightly rephrased now that we have a demonstration that complex numbers do provide an improvement over CNN. I gave it a go above}
%\sout{In addition to cyclic convolutions, the network operating in complex space could also contribute to this demonstrated robustness of LGCNs, as complex-valued neurons may provide better stability than real-valued neurons in CNNs and GCNs.}
%  \apn{a bit of redundancy to avoid confusion for fast readers}

\textbf{Prediction of cirrus structures in LSB images --} We task LGCNs with segmenting cirrus clouds in optical telescope images, demonstrating the effectiveness of our method on a real world problem. This dataset \cite{Sola2021CharacterizationImages} contains 48 expert annotated images of approximate size 5000x5000, with two channels representing different optical wavelength bands. Of the 48 images, we use 32 for training, 8 for validation (for 5-fold validation) and 8 for testing. Across the entire dataset 55\% of pixels are labelled as cirrus contaminated. In their original form, objects other than bright stars are indistinguishable from darkness, as the number of photons emitted by the brightest stars is several orders of magnitude larger than other objects such as cirrus clouds. To compensate for this we add an initial layer to the networks prior to projection onto complex space, implementing $\arcsinh$ scaling, popular in astronomical image processing, with learned parameters: $Y_{0}=\arcsinh{(aX_0+b)}$, where $a, b$ are real scalars. Following this a sigmoid function is applied and bad pixels are set to zero. Finally scaled images are concatenated onto original input to account for oversaturation of very strong cirrus regions. Initial scaling parameters were determined from a simple gradient descent algorithm, with the target set as auto-scaled versions of images using astronomical image software \cite{Boch2014AladinBrowser, Bonnarel2000TheSources} (GPL v3 license).

%\apn{Is there a justification for this change in architecture, or did you just find that this worked better? Did you also try it on the artificial images?} \farn{not entirely sure why I wrote this at the time - this network is the same as the above synthesised experiment} \apn{in that case, let's make sure that the architecture description is full and correct right from the start, and we can remove this part from here} 
%\sout{randomly selected cropped regions}
% \sout{through}
%\sout{ total samples}
Networks are trained over 300 epochs on random crops of size 512x512, which are then downscaled by a factor of two.
To mitigate against the limited sample size, we augment data with random flips and $90^\circ$ rotations, and pretrain networks on an extended version ($N$=1200) of the synthesised dataset.
We also train a standard CNN, a $\mathbb{C}$-CNN, and a GCN \cite{Luan2018GaborNetworks} for comparison as in the synthesised data experiment, fixing parameter size to be roughly equal. In comparison to the synthesised images, real cirrus regions often exhibit much fainter textures and orientation is more subtle and can vary slightly globally. In addition, training labels may not be fully reliable, due to the difficulty in annotating precisely the borders of cirri -- there is an inherent uncertainty associated with each annotation, especially due to the ambiguous nature of the cirrus cloud boundary, so several experts may disagree on the exact location of borders --, and due to the limited number of available expert annotators -- in our dataset, 2 to 3 annotators annotated each image, but for simplicity in this proof-of-concept, we worked only with the annotations of the single most experienced expert, making the simplified assumption that their annotation corresponds to the ground-truth. These factors, in combination with more severe artefacts, background noise, and small training size, make the dataset incredibly challenging. Results are shown in the last column of Table \ref{tab:denoising_results}: LGCNs achieve an IoU of 73.1\%, with an absolute increase of 4.7\% over standard CNN, 4.0\% over complex CNN and 4.6\% over real-valued static Gabor modulation without cyclic convolutions. Notably, GCN barely surpasses the base model and is outperformed by $\mathbb{C}$-CNN, suggesting that only static rotation sensitivity is not sufficient on more challenging datasets, a finding which is supported by results from the previous experiment of synthesised images. LGCNs significantly outperforms compared methods, demonstrating the ability of proposed augmentations to generate robust orientation sensitive features, even on data with extreme contamination. Given that the class balance is 55\%, the problem is very difficult, and although this absolute increase represents a significant performance improvement, more progress may be achieved by considering e.g. new architectures to be augmented by our methods, or a multi-scale approach, and a more complete dataset with consensus annotation from several experts.

\subsection{Application to boundary detection in natural images}
\label{subsec:BSD}

% \sout{by evaluating}
We demonstrate the general applicability of learnable Gabor modulated convolutions on the Berkeley Segmentation Dataset (BSD500) \cite{Arbelaez2011ContourSegmentation, Martin2001AStatistics}. This task requires the ability to learn equivariant features in the scenario where there is no dominant global orientation, and the network must handle high variations in local feature orientation dependence. The dataset contains natural images of size $321\times481$ in both portrait and landscape, with 200 training samples, 100 validation samples and 200 testing samples. Each image has associated with it several ground truth labellings produced by different annotators.

We replicate the pipeline of one of the highest performing methods, RCF \cite{Liu2017RicherDetection}, and replace convolutional operators with learnable Gabor modulated convolutions. This methodology uses a pretrained VGG16 \cite{Simonyan2015VeryRecognition} based architecture, taking 'side' outputs from each convolutional block that represent coarser scale edges as network depth increases. These side outputs are then fused together with a $1\times1$ convolutional layer. The final prediction is then computed as the average between all side outputs and the fused output. We denote our modified implementation as LGCN-RCF: an additional Gabor convolutional layer is used to create an orientation channel; all convolutional layers apart from the fusion layer are replaced with cyclic Gabor convolutions; orientation features are pooled prior to side output. We train for 250 epochs with 3-fold validation. Results are shown in Table \ref{tab:bsd}, using the optimal dataset scale (ODS) and optimal image scale (OIS) metrics defined in \cite{Arbelaez2011ContourSegmentation}. LGCN-RCF achieves 0.727 ODS and 0.747 OIS, which is a strong result considering in each epoch we train on one random augmentation per image, as opposed to other methods which use the entire range of augmentations per image (due to lack of compute and time, thus care is to be taken when comparing results).
Our method also significantly outperforms our implementation of RCF \cite{Liu2017RicherDetection} with parameters restricted to match LGCN-RCF, demonstrating the benefit of modulating with complex Gabor filters on tasks with natural images. 

\begin{table}
\small
\caption{Boundary detection results on the BSD500 \cite{Arbelaez2011ContourSegmentation} dataset. *Our parameter restricted implementation. $\dagger$ImageNet pretrained.}
\begin{tabular}{lllllll}
        \hline
          & \begin{tabular}[c]{@{}l@{}}Kivinen \\et al. \cite{Kivinen2014VisualDissection}\end{tabular}      & DexiNed \cite{Poma2020DenseDetection} & RCF* \cite{Liu2017RicherDetection} & H-Net \cite{Worrall} & LGCN-RCF & RCF$^\dagger$ \cite{Liu2017RicherDetection} \\ \hline
ODS       & 0.702 & 0.728   & 0.707 & 0.726 & 0.727         & 0.806         \\
OIS       & 0.715 & 0.745   & 0.720 & 0.742 & 0.747         & 0.823         \\ \hline
\# params & -     & 4.41M    & 1.80M  & 0.12M  & 1.88M     & 14.84M    \\ \hline    
\end{tabular}
\label{tab:bsd}
\end{table}

%---------------------------
\section{Conclusion} 
\label{sec:concl}

We presented a framework for incorporating adaptive modulation into complex-valued CNNs. This framework was used to design an orientation robust network with convolutional layers using Gabor modulated weights, where complex convolutional filters and Gabor parameters are learned simultaneously. A cyclic convolutional layer was proposed to retain rotational information throughout layers and encourage equivariance. Our architecture is able to generate unconstrained representations dependent on exact orientations, without interpolation artefacts. We validated this empirically for three use cases, with experiments designed to test properties of both invariance and equivariance to orientation. We first verified that LGCNs are able to effectively produce rotation invariant features on the rotated MNIST dataset. An ablation study was performed to assess in turn and in combination the effect of two proposed augmentations to GCNs \cite{Luan2018GaborNetworks}, namely using complex-valued weights and learning parameters of modulating Gabor filters. Secondly, we carried out experiments on a purpose designed dataset of varying difficulty. The architecture's modulated layers were able to create fine segmentations in synthetic and real images despite local disturbances. The presented LGCN architecture achieved strong denoising scores in comparison to standard CNNs, even on contaminating cirrus cloud structures with randomised orientation. Clear performance improvements were observed for both use cases, demonstrating the effectiveness of the augmentations. Thirdly, we applied an LGCN architecture to boundary detection in natural images and achieved strong metrics in comparison to other non-pretrained methods. The successful augmentation of three different architectures also demonstrates the general applicability of our method, and it may be applied to more complex DNNs and application scenarios in the future.
%\apn{an ablation study that also looks at the effect of the cyclic framework would be nice}

%-------------------
\section*{Funding}
This research did not receive any specific grant from funding agencies in the public, commercial, or not-for-profit sectors.

%-------------------
\bibliography{references}
\bibliographystyle{elsarticle-num-names}

\appendix
\begin{comment}
\section{Appendix}

Optionally include extra information (complete proofs, additional experiments and plots) in the appendix.
This section will often be part of the supplemental material.
\end{comment}

\end{document}